\pdfoutput=1
\documentclass[11pt]{article}
\usepackage{authblk}

\usepackage[]{acl}

\usepackage{times,xspace}
\usepackage{latexsym}

\usepackage{helvet} % DO NOT CHANGE THIS
\usepackage{courier}  % DO NOT CHANGE THIS
\usepackage{graphicx} % DO NOT CHANGE THIS
\usepackage{natbib}  % DO NOT CHANGE THIS AND DO NOT ADD ANY OPTIONS TO IT
\usepackage{caption} % DO NOT CHANGE THIS AND DO NOT ADD ANY OPTIONS TO IT

\usepackage{graphicx}
\usepackage{booktabs}
\usepackage{subfigure}
\usepackage{algorithm}
\usepackage{algorithmic}
\usepackage{color}

\usepackage{microtype}
\usepackage{amsmath, amssymb, amscd, amsfonts}
\usepackage{IEEEtrantools}
\usepackage{soul}
\usepackage{url}
\usepackage{algorithm}
\usepackage{algorithmic}
\usepackage{amssymb}
\usepackage{amsfonts} 
\usepackage{multirow}
\usepackage{boldline}
\usepackage{hhline}
\usepackage{xcolor}

\usepackage{makecell}
\usepackage{mathrsfs}
\usepackage[normalem]{ulem}
\useunder{\uline}{\ul}{}
\usepackage[inline]{enumitem}
\usepackage{enumitem}

\usepackage[T1]{fontenc}
\usepackage[utf8]{inputenc}

\title{A Multi-Format Transfer Learning Model for Event Argument Extraction via Variational Information Bottleneck}

\author[1$\dagger$*]{\bf Jie Zhou}
\author[2$\dagger$]{\bf Qi Zhang}
\author[2]{\bf Qin Chen}
\author[1]{ \bf Qi Zhang}
\author[2]{ \bf Liang He}
\author[1]{\bf Xuanjing Huang}
\affil[1]{School of Computer Science, Fudan Univerisity, Shanghai, China}
\affil[2]{School of Computer Science and Technology, East China Normal University, Shanghai, China}
\affil[  ]{\tt \{jie\_zhou, qz, xjhuang\}@fudan.edu.cn}
\affil[  ]{\tt qzhang@stu.ecnu.edu.cn; \{qchen, lhe\}@cs.ecnu.edu.cn}

\begin{document}
\maketitle

\renewcommand{\thefootnote}{\fnsymbol{footnote}}
\footnotetext[1]{Corresponding authors; $^\dagger$Equal contribution.}
\renewcommand{\thefootnote}{\arabic{footnote}}

\begin{abstract}

Event argument extraction (EAE) aims to extract arguments with given roles from texts, which have been widely studied in natural language processing. Most previous works have achieved good performance in specific EAE datasets with dedicated neural architectures. Whereas, these architectures are usually difficult to adapt to new datasets/scenarios with various annotation schemas or formats. Furthermore, they rely on large-scale labeled data for training, which is unavailable due to the high labelling cost in most cases. In this paper, we propose a multi-format transfer learning model with variational information bottleneck, which makes use of the information especially the common knowledge in existing datasets for EAE in new datasets. Specifically, we introduce a shared-specific prompt framework to learn both format-shared and format-specific knowledge from datasets with different formats. In order to further absorb the common knowledge for EAE and eliminate the irrelevant noise, we integrate variational information bottleneck into our architecture to 
refine the shared representation. We conduct extensive experiments on three benchmark datasets, and obtain new state-of-the-art performance on EAE.
% in both fully-supervised and low-data scenarios. 

%As an important subtask of event extraction (EE), event argument extraction (EAE) aims to extract the arguments assigned with roles. Due to the complexity of the EAE task, diverse datasets have various labeling spaces, formats and guidelines, which makes the transfer between different datasets difficult. In this paper, we propose a multi-format transfer learning model based on variational information bottleneck for EAE, denoted as \texttt{UnifiedEAE}. First, we introduce a shared-specific prompt framework to learn both format-shared and format-specific knowledge among different formats. Second, to enhance the model to eliminate useless information and retain the shared knowledge, we integrate variational information bottleneck into our architecture to learn the format-shared representation. We conduct extensive experiments on three benchmark datasets and obtain new state-of-the-art results over all the datasets. 
\end{abstract} 

% \ing cite{li2021invariant}
\section{Introduction}

Event Extraction (EE) has received widespread attention in recent years, which aims to obtain structured information (e.g., trigger, event types, arguments, and argument roles) from large unstructured text corpora \cite{lu2021text2event,zhang2022enhancing}. 
Event argument extraction (EAE) plays a crucial role in EE.
Recently, deep learning-based models have obtained tremendous success in this task.
However, these methods rely on a large-scale labeled dataset for training, which is time-consuming and labor-intensive due to the complexity of event extraction.

\begin{figure}[t!]
% \vspace{-1mm}
    \centering
    \includegraphics[scale=0.50]{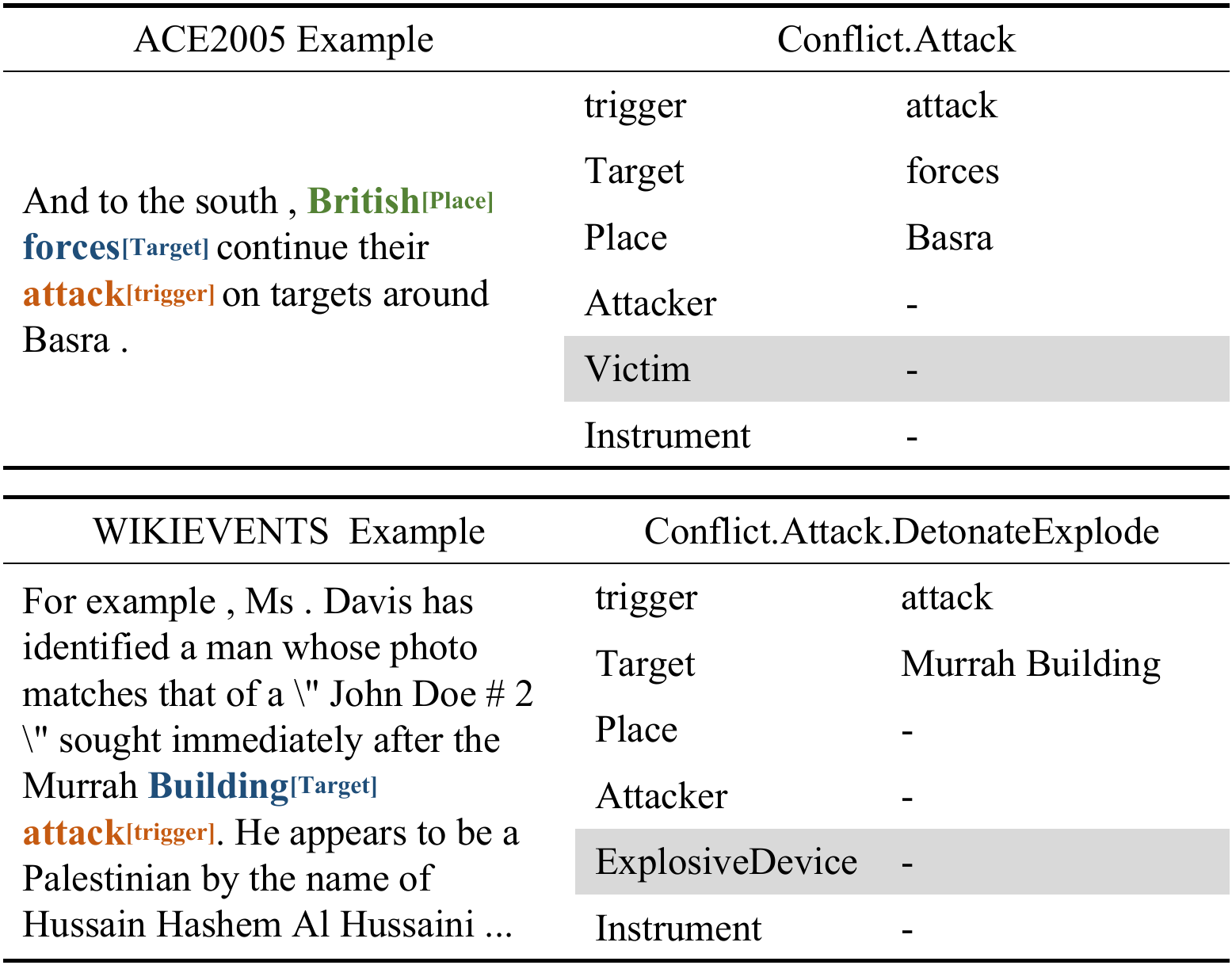}
    \caption{An example with a different format.}
    \label{fig:example}
    % \vspace{-2mm}
\end{figure}

In this paper, we aim to answer the question ``Can we transfer the knowledge from the existing complex event extraction datasets with different formats?". 
There are several event extraction datasets, such as ACE 2005 \cite{doddington2004automatic}, RAMS \cite{ebner-etal-2020-multi}, and WikiEvents \cite{li2021document}.
These datasets contain abundant event types and semantic roles that may possess overlap knowledge and help to improve the performance of new datasets or low-resource extraction. 
% For example, the ACE 2005 dataset \cite{doddington2004automatic} defines 33 different event types and 35 semantic roles. 
% RAMS \cite{ebner-etal-2020-multi} is a document-level dataset annotated with 139 event types and 65 semantic roles. In the WikiEvents dataset \cite{li2021document}, which collected 246 documents and annotated with 50 event types and 59 argument roles.
% WikiEvent
% RAMS
% MAVEN \cite{wang2020maven} contain 4,480 Wikipedia documents, 118, 732 event mention instances, and 168 event types.
% however it only apply to the task of event detection. 
As shown in Figure \ref{fig:example}, both ACE2005 and WikiEvents datasets contain the same ``attack" event type with inconsistent names.
Additionally, some shared argument roles (e.g., ``Target", ``Attacker", ``Place", and ``Instrument") are labeled in both two datasets.
All this information shows that the event knowledge can be transferred between two datasets.

% Moreover, there also have some Chinese dataset, DuEE-fin which contains 13 event types, 92 theoretical roles, and it has a labeled trigger word without specific positions, ChFinAnn \cite{zheng2019doc2edag} consisting of large scales of financial announcements and it only focus on five event types. Above all these datasets labeled the different event type and argument role.
%DUEE-Fin https://aistudio.baidu.com/aistudio/competition/detail/46/0/task-definition

% Recent successful approaches to event extraction have benefited from dense features extracted by neural models.\cite{chen2015event,nguyen2016joint,liu2018jointly} 
% Event extraction is challenging due to the complex structure of event records and the semantic gap between text and event. 
%  the varied of event labeling spaces, and the guide of annotation is also inconsistent. 
However, the transfer between different event argument extraction is a challenging task.
\textbf{(C1)} 
One challenge is that the formats of various datasets are inconsistent due to the complex structure of event records. 
Thus, it is hard to find a unified model to extract arguments with different formats.
More specifically, 1) Two datasets may have different event types, which have various argument structures;
2) The same event type or argument type in two datasets may have different names.
For example, the event names are ``Conflict Attack" and ``Conflict Attack Detonate Explode" in ACE2005 and WikiEvents respectively (Figure \ref{fig:example});
3) The argument roles set of the same event type may be different in various datasets. 
For instance, the argument role ``Victim" and ``ExplosiveDevice" for event ``Attack" only appear in ACE2005 and WikiEvents, respectively (Figure \ref{fig:example}).
\textbf{(C2)} The another challenge is that the annotation among different datasets may exist a gap, which brings noise for transfer learning. 
Two datasets may have significant semantic differences, as they may belong to different domains.
In addition, the annotation guidelines may be contradictory among various datasets.
% The annotation for the same argument role name may also be inconsistent since the guidelines for datasets are diversified. 
% In other words, one sentence may be labelled differently for the same argument role under different. 
Our experiments also show that merging two datasets simply may reduce the performance.

% However, the transfer between different event argument extraction is challenging due to the complex structure of event records, which makes the format inconsistent.
% \textbf{First}, the formats of various datasets are inconsistent.
% For, two different datasets may have different event types. 
% The structures for various event types may differ. 
% Also, one event type in two datasets may have different names.
% For example, the event names are ``Conflict Attack" and ``Conflict Attack Detonate Explode" in ACE2005 and WikiEvents respectively (Figure \ref{fig:example}).
% \textbf{Second}, the argument roles of the same event type may be different in various datasets. 
% Particularly, the roles set for the same event type may be different. 
% For instance, the argument role ``Victim" in ACE2005 is not in WikiEvents and ``ExplosiveDevice" in WikiEvents is not in ACE20005 (Figure \ref{fig:example}).
% Moreover, the same argument role in different datasets may have different names. 
% \textbf{Third}, the annotation for the same argument role name may also be inconsistent since the guidelines for datasets are diversified. 
% In other words, one sentence may be labelled differently for the same argument role under different annotation guidelines. 

Previous works mainly regard the argument extraction as a sequence labeling, which can not transfer to new event argument types \cite{yang2018dcfee}. 
Then, a machine reading comprehension problem (MRC) based model is proposed to extract the arguments using natural questions \cite{liu2020event,du2020event}.
Recently, prompt-learning \cite{schick2020exploiting,liu2021pre} based models \cite{ma2022prompt,chen2020reading} and generation-based models \cite{chen2020reading,du2021grit,li2021document} are utilized for event argument extraction. 
These studies inspire us to design a unified model that can extract arguments with different formats for EAE.
Moreover, some researches investigate cross-lingual event extraction \cite{subburathinam-etal-2019-cross} and zero-shot event extraction \cite{chen2020reading,feng2020probing}, which are under zero-shot setting.
In other words, these studies train on the source language or domain and transfer it to the target domain, where the target domain has no training data. 
Different from them, we train our model on both the source and target datasets with different formats where the format-shared knowledge is essential. 
% According to our previous observations, the datasets play an important role in training the event extraction model and we have a number of event datasets. So we face the question: "How to leverage all of the datasets?"

To deal with the above challenges, we propose a multi-format transfer learning model for EAE via information bottleneck,  denoted as \texttt{UnifiedEAE}, which can leverage all event extraction datasets with heterogeneous formats. 
First, we adopt a Shared-Specific Prompt (\texttt{SSP}) framework to capture format-shared and format-specific knowledge to extract arguments with different formats.
Then, to better capture the format-shared representation, we incorporate the variational information bottleneck (VIB) into the format-shared model (\texttt{SharedVIB}). 
VIB has been widely used to forget the irrelevant information and retain the vital information for prediction \cite{li2019specializing,tishby2000information}. 
We leverage it to enhance the model to learn the format-shared knowledge.
We conduct a series of experiments on three publicly available datasets and obtain new state-of-the-art performance. 
Our \texttt{UnifiedEAE} can also improve the performance of low-resource EAE effectively. 
Furthermore, the results show that our model can capture the format-shared knowledge and forget the noise among various datasets.

In summary, the main contributions of this paper are summarized as follows.
% \vspace{-1mm}
\begin{itemize}[leftmargin=*, align=left]
 \item We design a unified architecture that can learn both the format-shared and format-specific knowledge from various EE datasets with heterogeneous formats.
%  \vspace{-1mm}
 \item The information bottleneck technology is utilized to enhance the model to learn the format-shared knowledge among different datasets by eliminating the irrelevant information and reserving the format-shared knowledge. 
%  \vspace{-1mm}
 \item Extensive experiments on three datasets show the great advantages of our model. Also, our model performs well on low-resource event argument extraction.
\end{itemize}

\begin{figure*}[t!]
% \vspace{-7mm}
    \centering\includegraphics[scale=0.55]{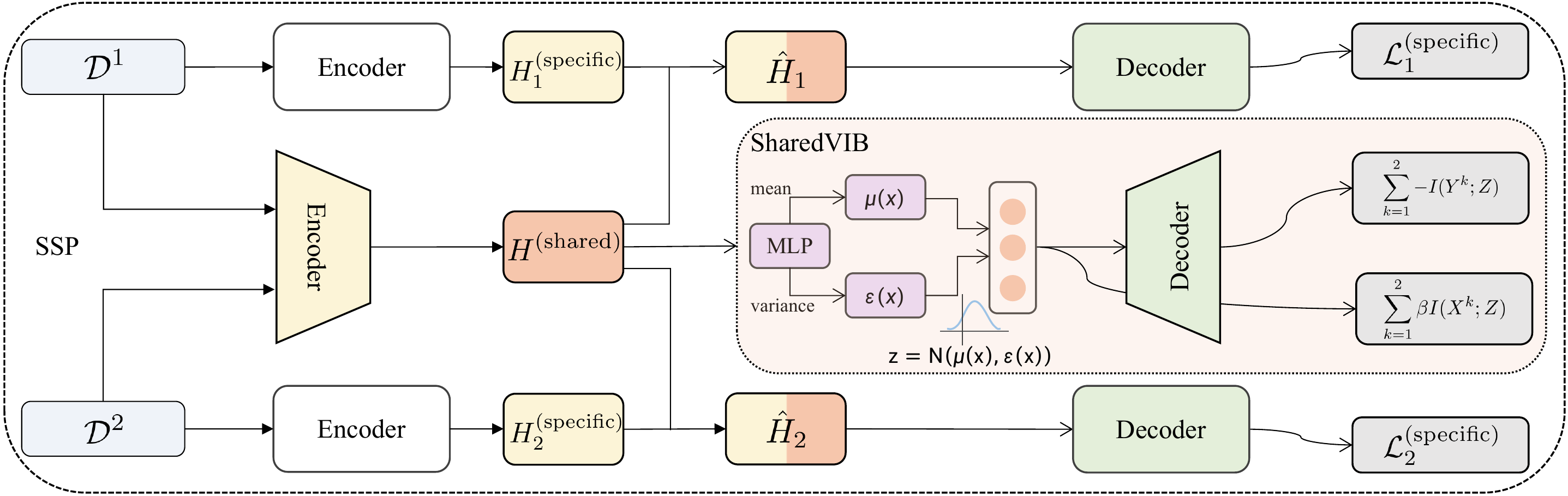}
    \caption{The framework of our \texttt{UnifiedEAE} model. To learn both the format-shared and format-specific representations (i.e., $\hat{H}_1$ and $\hat{H}_2$), we introduce a shared-specific prompt (\texttt{SSP}) model (white background). 
    Then, we design a \texttt{SharedVIB} module (pink background) to better capture the format-shared representation (e.g., $H^{\mathrm{(shared)}}$) by forgetting format-specific information ($\sum_{k=1}^{2}\beta I(X^k; Z)$) and retaining format-shared knowledge ($\sum_{k=1}^{2}-I(Y^k; Z)$) via information bottleneck.}
    \label{fig:framework}
    % \vspace{-3mm}
\end{figure*}

\section{Related Work}

\subsection{Event Argument Extraction}
Event extraction can be split into two subtasks, event identification and event argument extraction (EAE) \cite{zhang2020two,chen2015event,ijcai2022-589}.
We focus on the EAE task, which aims to extract the arguments based on the given event type and trigger \cite{wei2021trigger,ma2022prompt}.
\citet{wei2021trigger} added constraints
with each argument role to take the interaction into account. 
Data augmentation is adopted for event argument extraction \cite{liu2021machine}.
To avoid the error propagation and learn the relationships among the subtasks, end to end model performs two subtasks jointly \cite{zhang2019joint,wadden2019entity,li2021document}. 
Several studies regard event argument extraction as a machine reading comprehension problem (MRC), which extracts the arguments based on natural questions \cite{liu2020event,du2020event}.
Recently, prompt-learning \cite{schick2020exploiting,liu2021pre} based models \cite{ma2022prompt,chen2020reading} and generation-based models \cite{chen2020reading,du2021grit,li2021document} are utilized for event argument extraction.
In this paper, we aim to transfer the knowledge of the existing event extraction datasets to the target dataset, which is not well studied since the complexity of this task.

\subsection{Transfer Learning for NLP}
To reduce the requirements of labeled data, transfer learning has attached great attention in the field of natural language processing \cite{liu2017adversarial,ruder-etal-2019-transfer,raffel2020exploring,zhou-etal-2020-sentix}.
\citet{liu2017adversarial} proposed an adversarial multi-task learning framework to learn the shared and private representation.
Cross-lingual event extraction aims to transfer the knowledge from the source language to the target language \cite{subburathinam-etal-2019-cross}. 
Zero-shot transfer learning is also explored on semantic role labeling (SRL) \cite{peng2016event}, event extraction \cite{chen2020reading,feng2020probing}, and abstract meaning
representation (AMR) \cite{huang-etal-2018-zero}. 
Different from them, we focus on transfer learning among event argument extraction datasets with various complex formats where both the format-shared knowledge and format-specific knowledge are important.

\subsection{Information Bottleneck}
Recently, information bottleneck (IB) has been applied in NLP tasks, such as word cluster \cite{pereira1994distributional}, dependent parsing \cite{mahabadi2021variational}, summarization \cite{west2019bottlesum}, interpretability \cite{zhou2021attending}.
\citet{li2019specializing} use IB for compressing the hidden representations of words by removing the task-irrelevant information. 
\citet{sun2021graph} adopted the IB principle for graph structure learning. 
Variational IB (VIB) is used as a regularization technique to improve the fine-tuning of pre-training language models in low-resource scenarios \cite{mahabadi2021variational}.
In this paper, we attempt to use VIB to constraint model to learn format-shared information for event argument extraction.

\section{Methodology}
To transfer the knowledge among the datasets with different formats, we propose a \texttt{UnifiedEAE} model for event argument extraction task (Figure \ref{fig:framework}).
\texttt{UnifiedEAE} is based on a shared-specific prompt (\texttt{SSP}) architecture, which learns both the format-shared and format-specific knowledge from diverse datasets with multiple formats. 
% Also, this architecture can adapt to different formats of event arguments.
Then, to enhance the model to learn the format-shared knowledge, we integrate variational information bottleneck into the format-shared model (\texttt{SharedVIB}) by removing the format irrelevant information and retaining format invariable knowledge. 

Formally, given two event argument extraction datasets, denoted by $\mathcal{D}^1=\{(X^1_i, Y^1_i)\}^{|\mathcal{D}^1|}_{i=1}$ and $\mathcal{D}^2=\{(X^2_i, Y^2_i)\}^{|\mathcal{D}^2|}_{i=1}$ where $|\mathcal{D}^1|$ and $|\mathcal{D}^2|$ are the number of samples in dataset $\mathcal{D}^1$ and $\mathcal{D}^2$.
For each sample $(X, Y) \in \mathcal{D}^1 or \mathcal{D}^2$, the input $X=\{s, e, t, R\}$ contains the sentence $s$, event type $e$, and trigger word $t$, $R$ denotes the set of event-specific role types, we aims to extract a set of span $Y$. 
For the $i$-th span in $Y$, $\mathrm{span}^{\mathrm{start}}_i$ and  $\mathrm{span}^{\mathrm{end}}_i$ are the start and end indices of the span.

\subsection{Shared-Specific Prompt}
\label{sect:Shared-Specific Prompt Architecture}
The shared-specific prompt (\texttt{SSP}) architecture aims to learn both format-shared and format-specific knowledge for EAE. 
This framework consists of three event argument extractors: two format-specific and one format-shared extractor, which are used to learn the format-specific and format-shared knowledge.
We adopt a prompt-based model as the basic extractor to predict multi-format arguments.

\paragraph{Basic Prompt-based Extractor.} 
Following the \citet{ma2022prompt}, we use a BART \cite{lewis-etal-2020-bart} based prompt model as an event argument extractor.
This model consists of an encoder and decoder. 
The encoder is used to learn the event-aware sentence representation.
Then we adopt a decoder model to extract all the argument spans jointly via a prompt template.

\textbf{Encoder.} 
To consider the position information of event, 
we insert special token ``<t>" and ``</t>" before and after the trigger $t$ in the sentence $s$.
Then we input it into BART to obtain the event-aware sentence representation $H$,
\begin{equation}
\begin{aligned}
% \nonumber
\label{bartencoder}
       &  H_{\mathrm{encoder}} = \mathrm{BART_{Encoder}}(s), \\
       &    H = \mathrm{BART_{Decoder}}(s, H_{\mathrm{encoder}}),
\end{aligned}
\end{equation}

\textbf{Decoder.} In the decoder, we use a prompt with slots to extract the argument roles at the same time.
We use manual template from \citet{li2021document}. 
For example, the prompt is ``\underline{Person} married \underline{Person} at \underline{Place} ( and \underline{Place} )" for event type ``Life.Marry" in ACE2005. We aim to predict the argument spans for the four argument role slots.
We input the prompt $p$ to the BART decoder to obtain the prompt representation. 
\begin{equation}
\nonumber
    H_p = \mathrm{BART_{decoder}}(p, H_{\mathrm{encoder}})
\end{equation}

For the $i$-th role slot in the prompt, we use the mean pooling of the corresponding tokens' representation from $H_p$ as the role representation $r_i$. 
Then, we extract the argument span for role $r_i$ by predicting the start and end index in the text.
\begin{equation}
% \nonumber
\label{equ:classifier}
\begin{aligned}
p^{\mathrm{(start)}}_i & =\mathrm{Softmax}(r_{i}w^{\mathrm{(start)}}H) \\
p^{\mathrm{(end)}}_i & =\mathrm{Softmax}(r_{i}w^{\mathrm{(end)}}H)
\end{aligned}
\end{equation}
where $w^{\mathrm{(start)}}$ and $w^{\mathrm{(end)}}$ are the learnable parameters. 

Finally, the cross-entropy between the predicted start/end probability $p^{\mathrm{(start)}}_i$/$p^{\mathrm{(end)}}_i$ and the ground truth,
\begin{equation}
% \nonumber
\label{equ:baseloss}
\begin{split}
\mathcal{L} = \sum_{X \in D} \sum_{i=1}^{|R|}\mathrm{CrossEntropy}(p^{\mathrm{(start)}}_i, \mathrm{span}_i^{\mathrm{(start)}}) \\ + \mathrm{CrossEntropy}(p^{\mathrm{(end)}}_i, \mathrm{span}_i^{\mathrm{(end)}})
\end{split}
\end{equation}
where $\mathrm{span}_i^{\mathrm{start}}$/$\mathrm{span}_i^{\mathrm{end}}$ are the start and end index of the $i$-th argument role's span, $|R|$ is the length of roles set $R$ in $X$. 

For dataset $\mathcal{D}^1$ and $\mathcal{D}^2$, we use two independent prompt-based extractors to learn format-specific sentence representations $H^{\mathrm{(specific)}}_1$ and $H^{\mathrm{(specific)}}_2$ calculated by Equation \ref{bartencoder}. 
% \paragraph{Shared-Specific Architecture}
% To learn both the format-specific and format-shared knowledge, we design a shared-specific architecture based on the basic prompt model.
% Particularly, we utilize three basic prompt models. Two models are used to learn the format-specific knowledge for dataset one and dataset two.
Then, the third extractor is adopted to learn the format-shared sentence representation $H^{\mathrm{(shared)}}$  among two datasets. 
To predict the event argument based on both format-specific and format-shared knowledge, we combine specific representation $H^{\mathrm{(specific)}}_k, k \in \{1,2\}$ and shared representation $H^{\mathrm{(shared)}}$ via a gate mechanism \cite{hochreiter1997long}.
\begin{equation}
\nonumber
\begin{aligned}
g_k & = \sigma\left(W_{g_k} \cdot\left[H^{\mathrm{(specific)}}_k, H^{\mathrm{(shared)}}\right]+b_{g_k}\right) \\
    \hat{H}_k & = g_k * H^{\mathrm{(specific)}}_k + (1-g_k) * H^{\mathrm{(shared)}}
\end{aligned}
\end{equation}
where $W_{g_k}, b_{g_k}$ are the trainable parameters, $\sigma$ is a sigmoid active function. 

Then, we extract argument span by replacing $H$ in Equation \ref{equ:classifier} with $\hat{H}_k$.
In this way, we can predict the arguments based on both the format-specific and format-shared knowledge.
Then, we obtain the format-specific loss $\mathcal{L}_{\mathrm{SSP}} = \mathcal{L}^{\mathrm{(specific)}}_1 + \mathcal{L}^{\mathrm{(specific)}}_2$, where $\mathcal{L}^{\mathrm{(specific)}}_1$ and $\mathcal{L}^{\mathrm{(specific)}}_2$ are the loss for dataset $\mathcal{D}^1$ and $\mathcal{D}^2$.

\subsection{Shared Knowledge Learning via VIB}
We hope the shared model in shared-specific prompt architecture to learn the format-shared knowledge while forgetting the format-specific knowledge. 
However, we do not add objectives to enhance the model to do this.
Inspired by \cite{li2019specializing}, we integrate variational information bottleneck (VIB) into our shared model (\texttt{SharedVIB}) to capture the format-shared knowledge while eliminating the format-specific information.

Particularly, the information bottleneck aims to learn a compressed representation $Z$, which maximizes mutual information with output $Y$ and minimizes mutual information with input $X$.
In this paper, we tend to let $Z$ retain the information about $Y^{k}, k \in \{1,2\}$ and remove the irrelevant information in $X^{k}, k \in \{1,2\}$. 
\begin{equation}
\nonumber
    \sum_{k=1}^{2} \beta I(X^{k}; Z) - I(Y^{k}; Z)  
\end{equation}
The shared model performs both dataset $\mathcal{D}^1$ and $\mathcal{D}^2$ at the same time to learn the format-shared knowledge.
Then, we let the model to forget the format-specific information in $X^{k}, k \in \{1,2\}$ by minimizes mutual information between $Z$ and $X^{k}, k \in \{1,2\}$. 

It is challenging to compute the mutual information $I(Y^k; Z)$ and $I(X^k; Z)$ directly. 
The same as \cite{li2019specializing}, we use variational inference to compute a variational upper bound for $I(X^k; Z)$ as follow,
\begin{equation}
\nonumber
\small
\begin{aligned}
&\overbrace{\underset{x}{\mathbb{E}}\left[\underset{z \sim p(z \mid x)}{\mathbb{E}}\left[\log \frac{p(z \mid x)}{q(z)}\right]\right]}^{\text {upper bound }}-\overbrace{\underset{x}{\mathbb{E}}\left[\underset{t \sim p(z \mid x)}{\operatorname{\mathbb{E}}}\left[\log \frac{p(z \mid x)}{p(z)}\right]\right]}^{\mathrm{I}(X^{k} ; Z)} \\
&=\underset{x}{\mathbb{E}}\left[\mathrm{KL}\left(p(z) \| q(z)\right)\right] \geq 0
\end{aligned}
\end{equation}

We optimize the upper bounder of $I(X^k; Z)$ to minimize it.
We use reparameterzation method \cite{DBLP:journals/corr/KingmaW13} to sample $Z$ from the latent distribution according to $p(z|x)$,
\begin{equation}
\nonumber
% \begin{aligned}
p(z \mid x)=\mathcal{N}\left(z \mid f^{\mu}(x), f^{\Sigma}(x)\right)
% \end{aligned}
\end{equation}
where $f^{\mu}(x)=H^{(shared)}\cdot W^{\mu}$ and $f^{\Sigma}(x)=H^{(shared)}\cdot W^{\Sigma}$ are the mean and variance of the latent Gaussian distribution, $W^{\mu}$ and $W^{\Sigma}$ are the learnable parameters.
Thus, we estimate $I(X^k; Z)=\mathrm{KL}\left(p(z \mid x) \| q(z)\right)$.
For $q(z)$, we let it be a standard diagonal normal distribution.
% According to the definition, the mutual information between $Z$ and $X^{k}$ can be calculated as $\mathrm{I}(X^k ; Z)  \stackrel{\text { def }}{=} \mathbb{E}_{x, z}\left[\log \frac{p(z \mid x)}{p(z)}\right]  = \mathbb{E}_{x}\left[\mathbb{E}_{z \sim p_{\theta}(z \mid x)}\left[\log \frac{p_{\theta}(z \mid x)}{p(z)}\right]\right]$.
% To calculate $p(z)$, VIB
% replaces $p(z)$ with some variational distribution $q(z)$
% \begin{equation}
% \begin{aligned}
% &\mathrm{I}(X^{1,2} ; Z)  \stackrel{\text { def }}{=} \mathbb{E}_{x, z}\left[\log \frac{p_{\theta}(z \mid x)}{p_{\theta}(z)}\right] \\
% & = \mathbb{E}_{x}\left[\mathbb{E}_{z \sim p_{\theta}(z \mid x)}\left[\log \frac{p_{\theta}(z \mid x)}{p_{\theta}(z)}\right]\right]
% \end{aligned}
% \end{equation}

% \begin{equation}
% \mathrm{I}(Y^{k} ; Z) \stackrel{\text { def }}{=} \mathbb{E}_{y, z \sim p}\left[\log \frac{p(y \mid z)}{p(y)}\right]
% \end{equation}

For $I(Y^{k}; Z)$, we calculate the variational lower bound,
\begin{equation}
\nonumber
\begin{aligned}
&\overbrace{\underset{y, z \sim p}{\mathbb{E}}{\left[\log \frac{p(y \mid z)}{p(y)}\right]}}^{\mathrm{I}(Y^{k} ; Z)}-\overbrace{\underset{y, z \sim p}{\mathbb{E}}{\left[\log \frac{{\psi}(y \mid z)}{p(y)}\right]}}^{\text {lower bound }}\\
&=\underset{z \sim p}{\mathbb{E}}\left[\mathrm{KL}\left(p(y \mid z) \| {\psi}(y \mid z)\right)\right] \geq 0
\end{aligned}
\end{equation}

Here, we use the decoder model in our shared-specific prompt (Section \ref{sect:Shared-Specific Prompt Architecture}) as ${\psi}(y \mid z)$ by replacing $H$ in Equation \ref{equ:classifier} with the sampled $Z$. 
Thus, the loss for optimizing ${\psi}(y \mid z)$ on $D^{k}, k \in \{1,2\}$ is the same as Equation \ref{equ:baseloss} by replacing the sampled $Z$ with $H$ Equation \ref{equ:classifier}, denoted as $\mathcal{L}^{(shared)}$. 

Thus, the loss function for \texttt{SharedVIB} is, 
\begin{equation}
\nonumber
\small
\begin{aligned}
\mathcal{L}_{\mathrm{SharedVIB}} = \sum_{k=1}^{2}\left( \mathcal{L}^{\mathrm{(shared)}}_k 
+ \beta \sum_{X \in D^k}\mathrm{KL}\left(p(z \mid x) \| q(z)\right)\right)    
\end{aligned}
\end{equation}

Finally, the total loss for our \texttt{UnifiedEAE} is,
\begin{equation}
\nonumber
    \mathcal{L} = \mathcal{L}_{\mathrm{SSP}} + \mathcal{L}_{\mathrm{SharedVIB}}
\end{equation}

% In the inference phase, we use the format-specific model to extract the arguments.
% Our model can also perform zero-shot event argument extraction using the format-shared model.

\section{Experimental Setups}
% In this section, we first provide the description of the datasets (Section \ref{sect:datasets}) and evaluation metrics (Section \ref{sect:Evaluation Metric}). 
% Then, to evaluate the effectiveness of our model, we select several baselines for compression (Section \ref{sect:baselines}).
% Finally, we provide the implementation details for the sake of implementation (Section \ref{sect:implementation}).

\begin{table}[t!]
\caption{Data statistics of RAMS and WikiEvents}
\label{table: statistics of rams and wikievent}
\setlength{\tabcolsep}{1mm}{\begin{tabular}{lcccc}
\hlineB{4}
Dataset    & Split & \#Doc  & \#Event & \#Argument \\ \hline
\multirow{3}{*}{RAMS}       & Train & 3194   & 7329    & 17026      \\
           & Dev   & 399    & 924     & 2188       \\
           & Test  & 400    & 871     & 2023       \\ \hline
\multirow{3}{*}{WikiEvents}  & Train & 206    & 3241    & 4542       \\
           & Dev   & 20     & 345     & 428        \\
           & Test  & 20     & 365     & 556        \\
\hlineB{4}
\end{tabular}}
\end{table}

\begin{table}[t!]
\caption{Dataset statistics of ACE2005}
\label{table: statistics of ace2005}
\setlength{\tabcolsep}{1mm}{\begin{tabular}{lcccc}
\hlineB{4}
Dataset & Split & \#Sent & \#Event & \#Argument \\ \hline
\multirow{3}{*}{ACE2005} & Train & 17,172 & 4202    & 4859       \\
        & Dev   & 923    & 450     & 605        \\
        & Test  & 832    & 403     & 576        \\
\hlineB{4}
\end{tabular}}
\end{table}

\begin{table*}[t!]
% \vspace{-6mm}
\centering
\caption{The main results of event argument extraction. The best results are marked with \textbf{bold}.}
\label{table: argument-extraction}
\begin{tabular}{l|ccccccc}
\hlineB{4}
\multicolumn{1}{c|}{}                        & \multicolumn{2}{c}{ACE2005}                                                                     & \multicolumn{2}{c}{RAMS}                                                                        & \multicolumn{3}{c}{WikiEvents}                                                                                                      \\
\multicolumn{1}{c|}{\multirow{-2}{*}{}} & Args-I                                         & Args-C                                         & Args-I                                         & Args-C                                         & Args-I                                         & Args-C                                      & Head-C                               \\ \hline
FEAE                                         & -                                              & -                                              & 53.50                                           & 47.40                                           & -                                              & -                                           & -                                    \\
DocMRC                                       & -                                              & -                                              & -                                              & 45.70                                           & -                                              & 43.30                                        & -                                    \\
OneIE                                        & 65.90                                           & 59.20                                           & -                                              & -                                              & -                                              & -                                           & -                                    \\
EEQA                                         & 68.20                                           & 65.40                                           & 46.40                                           & 44.00                                             & 54.30                                           & 53.20                                        & 56.90                                 \\
BART-Gen                                     & 59.60                                           & 55.00                                             & 50.90                                           & 44.90                                           & 47.50                                           & 41.70                                        & 44.20                                 \\
EEQA-BART                                    & 69.60                                           & 67.70                                           & 49.40                                           & 46.30                                          & 60.30                                           & 57.10                                        & 61.40                                 \\
PAIE                                         & 73.60                                           & 69.80                                           & 54.70                                           & 49.50                                           & 68.90                                           & 63.40                                        & \textbf{66.50} \\ \hline
\texttt{UnifiedEAE}                                         & \textbf{76.06} & \textbf{71.85} & \textbf{55.46} & \textbf{49.94} & \textbf{69.84} & \textbf{64.00} & 66.30                                 \\
\texttt{UnifiedEAE} (Zero-shot)                              & 42.25                                          & 34.60                                           & 10.88                                            & 8.49                                           & 30.27                                          & 25.90                                        & 40.72  \\
\texttt{UnifiedEAE} (Single)                                         & 72.77 & 68.82 & 53.32 & 48.29 & 68.31 & 63.40 & 66.16  \\
\texttt{UnifiedEAE} (Multiple)                                 & 74.34                                          & 70.80                                           & 54.62                                          & 49.09                                          & 67.63                                          & 62.66                                       & 66.28                                \\ 
\hlineB{4}
\end{tabular}
% \vspace{-2mm}
\end{table*}

\subsection{Datasets} 
\label{sect:datasets}
Our experiments are conducted on the three widely-used datasets for event argument extraction task: ACE2005 \cite{doddington2004automatic}, RAMS \cite{ebner-etal-2020-multi} and WiKiEvents \cite{li2021document}.
The ACE2005 dataset is a sentence-level extraction dataset that defines 33 different event types and 35 semantic roles.
The split of training, validating, and testing sets is the same as \cite{wadden2019entity}. 
The RAMS dataset focuses on a document-level argument extraction task, including 139 event types and 65 semantic roles. 
The WikiEvents dataset is another document-level argument extraction
dataset, 246 documents are provided, with 50 event types and 59 argument roles. 
Our experiments are conducted under the annotations of their conventional arguments.
The statistics of the datasets are listed in Table \ref{table: statistics of rams and wikievent} and Table \ref{table: statistics of ace2005}.
% We also adopt precision (P), recall (R), and F1 score (F1) as evaluation metrics to evaluate the performance.
\subsection{Evaluation Metric} 
\label{sect:Evaluation Metric}
Following \citet{ma2022prompt}, we adopt two popular evaluation metrics. 
(1) Argument Identification F1 score (Arg-I): we consider an argument span is correctly identified when the predicted offset fits the golden-standard span.
(2) Argument Classification F1 score (Arg-C): if both the span and the argument role type are matched with the golden standard, we consider the argument is correctly classified.
For the WikiEvents dataset, we also additionally evaluate Argument Head F1 score (Head-C) that only considers the matching of the headword of an argument, the same as \cite{li2021document}.

\subsection{Baselines} 
\label{sect:baselines}
To investigate the effectiveness of our model, we
compare our approach with the following state-of-the-art baseline models.

\begin{itemize}[leftmargin=*, align=left]
    \item ONEIE \cite{lin2020joint} is a joint neural model to extract the globally optimal IE result.
    \item BART-Gen \cite{li2021document} proposes a document-level neural event argument extraction model by regarding this task as conditional generation based on event templates.
    \item EEQA \cite{du2020event} proposes an end-to-end model and translates event extraction task into a question answering (QA) task.
    \item FFAE \cite{wei2021trigger} constructs the EAE task as a QA-based algorithm and uses the intra-event argument interaction to improve the performance.
    \item DocMRC \cite{liu2021machine} utilizes implicit knowledge transfer and explicit data augmentation based on a QA-based method.
    \item EEQA-BART \cite{ma2022prompt} replaces the BERT with BART for event extraction.
    \item PAIE \cite{ma2022prompt} utilizes prompt tuning for extracting argument extraction so that it can take the best advantages of pre-trained language models. 
    \item \texttt{UnifiedEAE} is our full model, which trains on two datasets and transfers to one of them. For \texttt{UnifiedEAE} (Zero-shot), we train our model on two datasets and test on the third dataset via format-shared extractor. 
    \item \texttt{UnifiedEAE} (Multiple) trains on the merged dataset, which removes the \texttt{SSP} from our \texttt{UnifiedEAE} model. In other words, it is a basic prompt-based extractor in Section \ref{sect:Shared-Specific Prompt Architecture}.
    Different from \texttt{UnifiedEAE} (Multiple), \texttt{UnifiedEAE} (Single) trains and tests on the same dataset without transferring.
\end{itemize}

\subsection{Implementation Details}
\label{sect:implementation}
We initialize the weight in encoder-decoder architecture
with pre-trained BART base models~\cite{lewis-etal-2020-bart}. 
We use Adam optimizer with the learning rates of 2e-5. 
The max encoder sequence length is 500, and the max decoder length is 80. 
The dropout is 0.1. 
The reported results on the test set are based on the parameters that obtain the best performance on the development set.

% \usepackage{multirow}

\section{Results and Analyses}
To investigate the efficacy of \texttt{UnifiedEAE} model, we compare our model with the mainstream baselines (Section \ref{sect:main results}).
Then, we do the ablation studies to verify the performance of the parts consisting of our model from two views, the model structure and dataset transferring (Section \ref{sect:ablation studies}). 
We also explore the effectiveness of transfer learning on low-resource EAE (Section \ref{sect:zero-shot and few-shot}) and case studies are given (Section \ref{sect:casestudies}). 

\subsection{Main Results}
\label{sect:main results}
% The best results of transfer learning 
In this section, we compare our framework with several prior competitive baselines (Table \ref{table: argument-extraction}). 
Note, for \texttt{UnifiedEAE} and \texttt{UnifiedEAE} (w/o \texttt{SSP}), we report the best results of transferring over each two datasets (e.g., ACE2005 and RAMS, ACE2005 and WikiEvents, RAMS and WikiEvents). 

From the table, we find the following observations.
\textbf{First}, we observe that our model consistently outperforms the state-of-the-art baseline in terms of Args-I and Args-C. 
Compared with the best baseline PAIE, our approach outperforms it with an improvement of 2.46\% in terms of Args-I over ACE2005, which indicates the effectiveness of multi-format transfer learning.
\textbf{Second}, our model can leverage the knowledge from other datasets effectively.
Our \texttt{UnifiedEAE} model outperforms \texttt{UnifiedEAE} (w/o \texttt{SSP}), which trains a basic prompt-based extractor using merged data directly. 
Moreover, \texttt{UnifiedEAE} (Single) sometimes performs better than \texttt{UnifiedEAE} (w/o \texttt{SSP}).
These indicate that merging two datasets directly for training may bring noise and lead to small gains or even drops.
% For PAIE (merge), the improvement is more modest or even decreased. Because the two datasets are directly merged for training, it may bring noise and lead to smaller gains. On the contrary, we achieve the best performance on the three datasets in most cases. 
% Second, our approach that integrates information bottleneck in our model exceeds the EEQA-BART baseline with a large magnitude, indicating that we can learn the common information across multiple data sets well. Third, 
\textbf{Third}, we apply our transfer learning framework to implement zero-shot. 
It is capable of extracting new event argument roles that are unseen in the training phase using the format-shared model.
From the results, we can observe that transferring between ACE2005 and WikiEvents achieves a good performance because they have many similar event types and argument roles.

% to answer the question
% \usepackage{multirow}
\begin{table}[!t]
\centering
\scriptsize
\caption{The performance of transfer learning between RAMS and WikiEvents.}
\label{table:transfer rams and wikievents}
\setlength{\tabcolsep}{1.5mm}{\begin{tabular}{l|ccccc}
\hlineB{4}
\multirow{2}{*}{} & \multicolumn{2}{c}{RAMS} & \multicolumn{3}{c}{WikiEvents} \\
                       & Args-I      & Args-C     & Args-I   & Args-C   & Head-C   \\ \hline
\texttt{UnifiedEAE}                    & \textbf{55.05}       & {49.71}      & \textbf{67.68}    & \textbf{62.79}    & \textbf{68.26}    \\
w/o \texttt{SharedVIB}                & 54.65       & 48.79      & 65.90     & 61.05    & 67.42    \\
w/o \texttt{SSP}                 & 54.87       & \textbf{49.92}      & 63.60     & 59.27    & 67.41   \\ \hline
\texttt{UnifiedEAE} (Single)                &  53.32  & 48.29    & 68.31 & 63.40     & 66.16  \\
\hlineB{4}
\end{tabular}}
\end{table}

% \usepackage{multirow}
\begin{table}[!t]
\centering
\scriptsize
\caption{The performance of transfer learning between ACE2005 and WikiEvents.}
\label{table:transfer ace and wikievents}
\setlength{\tabcolsep}{1.5mm}{\begin{tabular}{l|ccccc}
\hlineB{4}
\multirow{2}{*}{} & \multicolumn{2}{c}{ACE2005} & \multicolumn{3}{c}{WikiEvents} \\
                       & Args-I       & Args-C       & Args-I   & Args-C   & Head-C   \\ \hline
\texttt{UnifiedEAE}                    & \textbf{76.06}        & \textbf{71.85}        & \textbf{69.84}    & \textbf{64.00}       & 66.30     \\
w/o \texttt{SharedVIB}                & 75.12        & 71.48        & 68.76    & 63.27    & \textbf{67.76}    \\
w/o \texttt{SSP}                 & 74.34        & 70.80        & 67.63    & 62.66    & 66.28   \\
\hline
\texttt{UnifiedEAE} (Single)                & 72.77       & 68.82     & 68.31   & 63.40    & 66.16  \\
\hlineB{4}
\end{tabular}}
\end{table}

% \usepackage{multirow}
\begin{table}[!t]
\centering
\small
\caption{The performance of transfer learning between ACE2005 and RAMS.}
\label{table:transfer ace and rams}
\setlength{\tabcolsep}{0.8mm}{\begin{tabular}{l|cccc}
\hlineB{4}
\multirow{2}{*}{} & \multicolumn{2}{c}{ACE2005} & \multicolumn{2}{c}{RAMS} \\ 
                       & Args-I       & Args-C       & Args-I      & Args-C     \\ \hline
\texttt{UnifiedEAE}                   & \textbf{71.65}        & \textbf{68.00}        & \textbf{55.46}       & 49.94      \\
w/o \texttt{SharedVIB}                & 67.59        & 62.96        & 55.12       & \textbf{50.02}      \\
w/o \texttt{SSP}                 & 62.61        & 59.62        & 54.62       & 49.09      \\
\hline
\texttt{UnifiedEAE} (Single)      & 72.77       & 68.82     &  53.32 & 48.29   \\
\hlineB{4}
\end{tabular}}
\end{table}

\begin{figure*}[t!]
% \vspace{-4mm}
    \centering
    \includegraphics[scale=0.55]{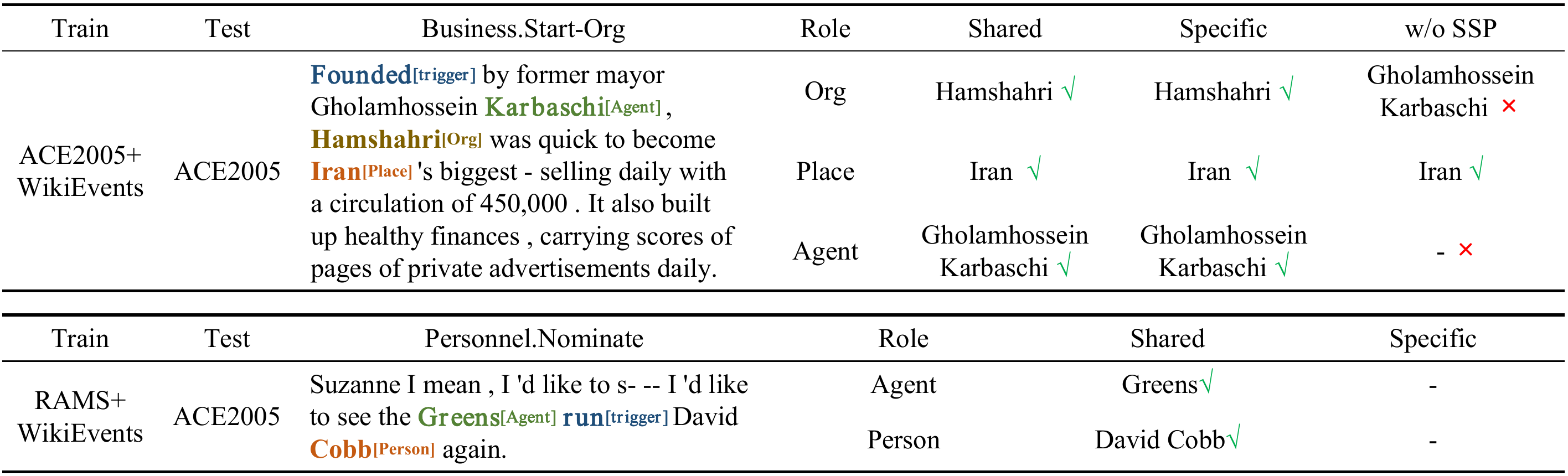}
    \caption{Examples Visualization. We show the results of format-shared and format-specific extractors.}
    \label{fig:case study}
    % \vspace{-2mm}
\end{figure*}

\begin{figure}[!t]
    \centering
    \subfigure[Transfer from WikiEvents to RAMS]{
    \label{RAMS-span-C}
    \begin{minipage}[t]{0.46\linewidth}
    \centering
    \includegraphics[width=1.47in]{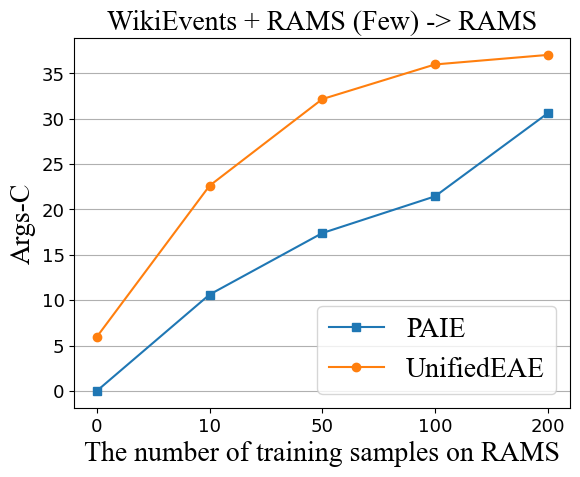}
    \end{minipage}
    }
    \subfigure[Transfer from WikiEvents to RAMS]{
    \label{pt-bert}
    \begin{minipage}[t]{0.46\linewidth}
    \includegraphics[width=1.47in]{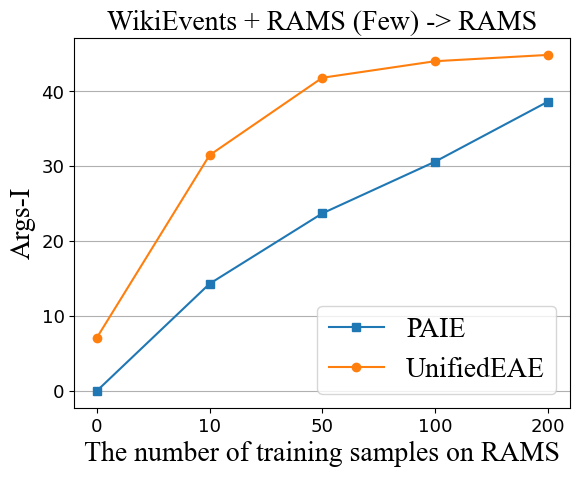}
    \end{minipage}
    }
    
    \subfigure[Transfer from ACE2005 to WikiEvents]{
    \label{scd-bert}
    \begin{minipage}[t]{0.46\linewidth}
    \includegraphics[width=1.47in]{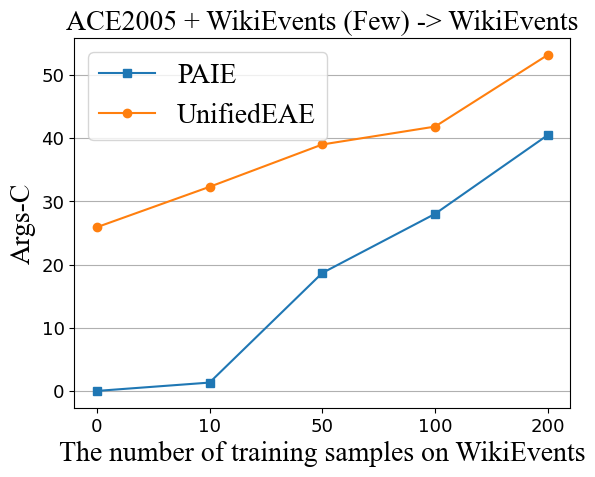}
    \end{minipage}
    }
    \subfigure[Transfer from ACE2005 to WikiEvents]{
    \label{informin-cl}
    \begin{minipage}[t]{0.45\linewidth}
    \includegraphics[width=1.47in]{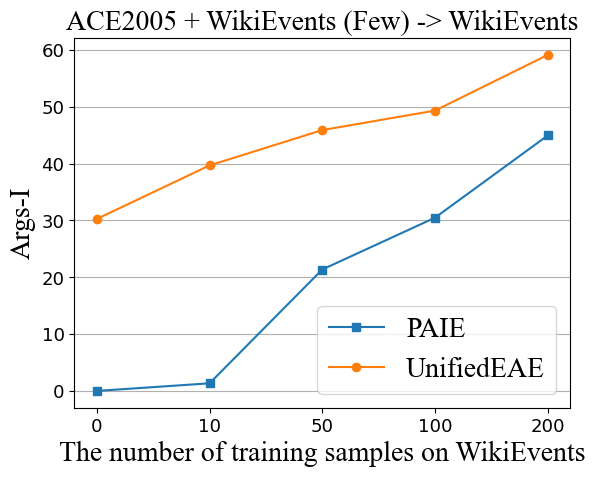}
    \end{minipage}
    }
    \caption{The results of low-resource event argument extraction via transfer learning.}
    \label{fig:low-resource}
\end{figure}

\subsection{Ablation Studies}
\label{sect:ablation studies}
We do the ablation studies to further investigate the effectiveness of the main components in our model from two perspectives, model structure and resource information. 
The results are shown in Table \ref{table:transfer rams and wikievents}, \ref{table:transfer ace and wikievents} and \ref{table:transfer ace and rams}. 
% To further investigate the effectiveness of our main components contained, we do ablation studies in this section.

From a model structure view, we remove the \texttt{SharedVIB} (w/o \texttt{SharedVIB}) and \texttt{SSP} (w/o \texttt{SSP}) from our model respectively. 
We observe that each component can help boost the performance of EAE. 
Particularly, \textbf{1)} Removing \texttt{SSP} will cause about four points decline in terms of argument identification and classification when 
transferring the knowledge of RAMS to WikiEvents (Row 1,3 in Table \ref{table:transfer rams and wikievents}).
This justifies that directly merging datasets may bring noises, which results in the degradation of test data.
Our model can learn both the format-shared and format-specific knowledge, which improves the performance effectively.
\textbf{2)} Our \texttt{SharedVIB} strategy can enhance the model to learn the format-shared knowledge.
Removing \texttt{SharedVIB} from our model will reduce the performance in most cases.
For example, \texttt{UnifiedEAE} obtains more than 4 points improvement compared with the one without \texttt{SharedVIB} when transferring RAMS to ACE2005.
% (2) The WikiEvents and RAMS shared knowledge brings as large as 2.3 absolute points for argument identification and 1.78 points for argument classification. (3) The shared-private further improves Arg-C scores in ACE2005 and WikiEvents, since the two datasets have more same argument role types.

%while taking a slightly negative effect on ACE05. Since the former two datasets are document-level and have more role types

To investigate the effectiveness of transfer learning among different resources, we evaluate our model on each two datasets. 
As we mentioned above, \texttt{UnifiedEAE} (Single) trains and tests on the same dataset and \texttt{UnifiedEAE} without \texttt{SSP} trains on the merged datasets.
We can find that not all the transferring can improve the performance since it may contain noise for the target dataset. 
For example, transferring from RAMS to ACE2005 caused more than six points drop for both Args-I and Args-C compared with \texttt{UnifiedEAE} (Single) that only trains on ACE2005 (Row 3 and 4 in Table \ref{table:transfer ace and rams}). 
Our model can reduce the influence of noise effectively by learning both format-shared and format-specific knowledge.

% \paragraph{The Effectiveness of Transfer Learning}
% \label{sect:The Effectiveness of Transfer Learning}
% Transfer learning for (ace, wiki) -> (ace, wiki)

% \begin{figure}[t]
% % \vspace{-4mm}
% \tiny
% \centering
% \includegraphics[width=0.49\columnwidth]{RAMS-span-Classification.png}
% \includegraphics[width=0.49\columnwidth]{RAMS-span-Identification.png}
% \includegraphics[width=0.49\columnwidth]{wiki-span-Classification.png}
% \includegraphics[width=0.49\columnwidth]{wiki-span-Identification.png}
% % \vspace{-1mm}
% \caption{Results of human evaluation.}
% \label{fig:zero-shot}
% \end{figure}

\subsection{Low-Resource EAE}
\label{sect:zero-shot and few-shot}
% The results of Shared (ace, wiki) -> (rams)
Furthermore, we explore the performance of low-resource event argument extraction via transfer learning (Figure \ref{fig:low-resource}).
We transfer the knowledge from the source dataset (e.g., WikiEvents) to the target dataset (e.g., RAMS) with few samples in the target dataset.
In our experiments, we train our model with 0, 10, 50, 100, and 200 samples.
From the results, we obtain the following observations.
First, \texttt{UnifiedEAE} significantly outperforms the state-of-the-art PAIE model in terms of both Args-C and Args-I over two datasets.
Particularly, our model achieves over 30 points with only ten examples on WikiEvents in terms of F1, while the PAIE model is almost not working.
Second, \texttt{UnifiedEAE} obtains better performance with fewer samples compared with PAIE. 
\texttt{UnifiedEAE} uses ten examples and performs even better than PAIE with 100 examples.
Third, \texttt{UnifiedEAE} with 200 samples achieves the comparable results with the existing baselines (e.g., BART-Gen, EEQA-BART) trained on full training datasets (3241 samples) on WikiEvents.
All these findings indicate that our model captures the format-shared and format-specific knowledge and transfers the format-shared information effectively.

\subsection{Case Studies}
\label{sect:casestudies}
To make it easier to understand how our \texttt{UnifiedEAE} model works, we visualize two examples on ACE2005 in Figure \ref{fig:case study}.
We find that our \texttt{UnifiedEAE} model transfers the knowledge effectively.
1) The format-shared module extracts the arguments correctly by learning the overlapping knowledge among multiple formats' datasets.  
However, \texttt{UnifiedEAE} (w/o \texttt{SSP}), which trains on the merging dataset directly, can not predict ``Org" and ``Agent".
2) We also train our model on RAMS and WikiEvents and test it on ACE2005 using a format-shared extractor, which is under a zero-shot setting.
We find our format-shared model can extract the event roles for the event ``Personnel.Nominate" though it does not appear in the training dataset. 
% We train our model on ACE2005 and WikiEvents and test on ACE2005.
% The reported results are based on the format-shared and format-specific modules.
% To better understand the mechanism by which \texttt{UnifiedEAE} improved performance, we show two typical examples of experiments that merge ACE2005 and WikiEvents datasets in Figure \ref{fig:case study}. 

\section{Conclusions and Future Work}
In this paper, we propose a unified event argument extraction (\texttt{UnifiedEAE}) model to transfer the knowledge among multi-format datasets.
First, a shared-specific prompt architecture is introduced to extract the event arguments with multiple formats based on both format-shared and format-specific representations. Then, to enhance the model to capture the format-shared knowledge effectively, we integrate the information bottleneck into our architecture.
Variational information bottleneck is leveraged to eliminate the format-specific information and retain the format-shared knowledge.
We conduct extensive experiments on three EAE datasets and compare our model with several strong baselines.
The results show that our \texttt{UnifiedEAE} model outperforms the state-of-the-art baselines.
Furthermore, the ablation studies show \texttt{SharedVIB} can capture the format-shared effectively.
Our model also obtains good results on low-resource event argument extraction. 
In further work, we would like to adopt our model to other complex tasks, such as relation extraction, and named entity recognition.

\section*{Acknowledge}
The authors wish to thank the reviewers for their helpful comments and suggestions. This work was partially National Natural Science Foundation of China (No. 61976056, 62076069), Shanghai Municipal Science and Technology Major Project (No.2021SHZDZX0103).
% This research is funded by the Science and Technology Commission of Shanghai Municipality (No. 19511120200\&21511100100 and 21511100402) and by Shanghai Key Laboratory of Multidimensional Information Processing, East China Normal University, No. 2020KEY001 and the Fundamental Research Funds for the Central Universities. 
% This research is also funded by the National Key Research and Development Program of China (No. 2021ZD0114002), the National Nature Science Foundation of China (No. 61906045), and Shanghai Science and Technology Innovation Action Plan International Cooperation project ``Research on international multi language online learning platform and key technologies (No.20510780100)". The computation is performed in ECNU Multi-functional Platform for Innovation (001).

\bibliography{custom}
\bibliographystyle{acl_natbib}

% \appendix

\end{document}